# ChatGPT is on the Horizon: Could a Large Language Model be Suitable for Intelligent Traffic Safety Research and Applications?

**Ou Zheng[1], Ph.D.**
Email: ou.zheng@ucf.edu

**Mohamed Abdel-Aty[1], Ph.D.**
Email: m.aty@ucf.edu

**Dongdong Wang[1], Ph.D.**
Email: dongdong.wang@ucf.edu

**Zijin Wang[1]**
Email: zijin.wang@ucf.edu

**Shengxuan Ding[1]**
Email: shengxuan.ding@ucf.edu

1 Department of Civil, Environmental, and Construction Engineering, University of Central Florida, Orlando, FL, 32816, USA

## ABSTRACT

ChatGPT embarks on a new era of artificial intelligence and will revolutionize the way we approach intelligent traffic safety systems. This paper begins with a brief introduction about the development of large language models (LLMs). Next, we exemplify using ChatGPT to address key traffic safety issues. Furthermore, we discuss the controversies surrounding LLMs, raise critical questions for their deployment, and provide our solutions. Moreover, we propose an idea of multi-modality representation learning for smarter traffic safety decision-making and open more questions for application improvement. We believe that LLM will both shape and potentially facilitate components of traffic safety research.

## INTRODUCTION

Intelligent transportation system is already beginning to revolutionize various aspects of our daily life, spanning from public transportation and urban planning to traffic safety decision-making. Among these innovative applications, intelligent traffic safety draws more attention since it is critical to the complex interaction between vulnerable road users and motor vehicles. Especially, industries and governments are promoting autonomous driving which has raised the concerns about the traffic safety for a mixed and complex traffic system. Therefore, increasing research on intelligent traffic safety explores more effective and efficient solutions with other emerging technologies. In recent decades, artificial intelligence (AI) technology, particularly deep learning, has advanced rapidly, achieving significant commercial achievements and public recognition on many fields, like computer vision (CV) and natural language processing (NLP). These achievements also provide new solutions to traffic safety problems [1]. For example, lots of accurate and efficient deep recognition models facilitate traffic condition analysis and road user behavior prediction [2]. Also, some powerful language models significantly increase textual processing and analysis, like accident report summarization and generation [3]. Moreover, with further advance in training techniques, the deep models are scaled up with massive amounts of data and one of these milestone works is large language models (LLMs), such as Generative Pre-trained Transformer (GPT) family. ChatGPT, as a member of GPT family, has demonstrated impressive language understanding and generation competence and strong capability to perform multi-domain tasks without fine-tuning. It significantly outperforms on various NLP challenging tasks, such as question answering and summarization, etc. Besides remarkable achievements in linguistic tasks, its potentials can be enhanced by incorporating cross-modality model from other data sources, such as sensors, cameras, and other Internet of Things (IoT) devices [4]. For example, accident detection and analysis can be simultaneously accomplished with accident scene understanding by integrating video recognition model with ChatGPT. This implies that the LLM, as an emerging technology, can facilitate traffic safety analysis and give rise to more efficient solutions.

We attempt to discuss the applications of LLM on intelligent traffic safety and share our perspective on their potential, opportunity, and threat. This work, other than proposes a specific solution, is intended to discuss LLM-integrated methods and LLM-equipped applications from a forward-looking and speculative viewpoint. First, we quickly walk through the development of LLM. Next, we focus the discussion on applying LLM to several typical traffic safety problems, including accident report automation [5], traffic data augmentation [6], and multisensory data analysis [7]. Furthermore, we discuss the risks on the application of LLM for traffic safety from the perspectives of model bias, data privacy, and artificial hallucination, and provide possible solutions to these problems. Moreover, inspired from success of LLM pretraining, we bring forward multi-modality representation modeling for intelligent traffic safety in future, which also brings both opportunities and challenges.

## DEVELOPMENT OF LLM

Language modeling is an important methodology to quantitatively analyze and predict the sequences of words with statistical and probabilistic models, which is widely adopted in NLP. Before the rapid development of LLM, statistical language models play a prominent role in NLP tasks due to its excellent performance by deductive reasoning, such as n-grams [8] and rule [9]. Recently, thanks to data explosion, deep learning optimization, and hardware supports, more complex and sophisticated machine learning language models are derived from observations in an inductive reasoning manner, like word2vec [10] and GloVe [11]. These models significantly outperform conventional statistical language models and direct increasing research focus on a large set of text corpus to model the semantic relationship between words. AllenNLP proposed ELMo [12] based upon Bi-LSTM, one of the earliest popular language models for representation learning. Its important success of language model scale-up inspires the research in large-scale language representation learning since then. In 2018, Google AI developed the first large pre-trained model, BERT [13], with masked modeling. BERT provides a more comprehensive and accurate semantic representation on words and can be integrated with other learning models to achieve huge success in a variety of NLP tasks. This model marks a notable milestone in the language modeling transition to large-scale representation learning. OpenAI built a transformer-based pre-training framework of GPT-1 [14], which significantly improves context understanding, especially text generation, through representation learning.

**Rising of Large-scale Pre-training.**

Given the success of BERT and GPT-1, the research teams in industry started exploring more advanced and sophisticated solutions to word representation by scaling up model architectures. Google proposed T5 [15], an important LLM with 11 billion parameters, which is trained by unifying all NLP tasks in text-to-text format. T5 attained huge success in the strong generalization to multiple downstream tasks, even without fine-tuning. This feature implies that the increase in training scale for comprehensive representation is more important than multiple stage tunings for multi-task solving. In the same year, OpenAI proposed GPT-3 [16] by increasing the number of model parameters to billions, which further significantly improved the representation and model generalization to diverse downstream tasks. Derived from this modeling framework, OpenAI developed InstructGPT [17], which incorporates reinforcement learning into instruction understanding. With further improvement, ChatGPT, a variant of InstructGPT, is fine-tuned in a more conversational manner and can more efficiently interact with humans. As another milestone, ChatGPT impressed many users by virtue of its exceptionally superior performance. It stirred up lots of attention from extensive fields due to its impressive analysis skills. ChatGPT, with billions of parameters, follows the prototype of the GPT family but is trained in an interactive manner through reinforcement learning. Given sophisticated reward model design and effective optimization over multiple expert models, ChatGPT exhibits more human-friendly nature and accurate reasoning. More importantly, large-scale training data enhances the knowledge base and enable ChatGPT to tackle more challenging semantic tasks.

**LLM Beyond Stand-alone Application.**

Although the LLMs like ChatGPT show impressive analysis skills, their competencies can be further empowered by integrating with other powerful models. For example, more deep learning research scientists explored cross-modal learning and attempt to unleash the power of vision-language modeling on image-to-text/text-to-image tasks. GAN [18], as a useful image generation prototype model, was integrated with text representation to generate various realistic images given narrative requirements. Recently, Stable Diffusion [19], based upon a latent diffusion model, achieved competitive performance on multiple generative tasks in a more efficient way. Google proposed Imagen [20] with text embedding from T5 for better context understanding and image generation quality. Inspired by language representation learning, more researchers also explored a variety of vision-language pre-training paradigms and developed a range of powerful large pre-trained models with strong domain transfer ability. OpenAI developed the model of CLIP [21] with contrastive learning to connect text and image for large-scale representation learning. Salesforce improved CLIP with BLIP [22] by leveraging unlabeled data from the internet as noisy data to improve representation generalization. In conjunction with CLIP, OpenAI designed DALL-E [23] to enhance the liaison between image and text by incorporating a variant of GPT-3 into image generation for more realistic results. To further improve image generation, DALL-E v2 [24] was developed by integrating diffusion models [19].

## THE APPLICATIONS OF LLM IN TRAFFIC SAFETY

In recent months, a range of successful application with ChatGPT cases have been demonstrated in intelligent transportation. However, focusing on traffic safety, few works have sought to explore the implementation of innovative LLM-based tools to facilitate traffic safety analysis. To fill this study gap, we demonstrate the potential applications on traffic safety analysis and decision-making in this paper. The discussion includes accident report automation, traffic data augmentation, and multisensory safety analysis.

### Accident Report Automation.

An accident report plays a central role in traffic safety analysis and evaluation, which shares important incidence facts with the stakeholders from different fields. Much important traffic safety research relies on the data from the report, such as Lee et al.; Farhangfar et al. [25, 26]. The information from accident reports records the details of a car accident and is usually prepared by transportation officers following structural format. However, when a car accident occurs, the persons witnessing or involved in the accident usually provide the narratives of the accident process, also the officer might write a description of the accident occurrence and draw a sketch of the accident scene, which hinders efficient quantitative analysis. The descriptions and records are further processed, formatted, and documented by human curation for future usage by stakeholders. This process may take much time for document processing. Human written materials can also result in bias problems due to the uncertainty between different writers. The early language models can tackle this problem; however, they still exhibit moderate levels of accuracy and reliability. By training with enormous amounts of data, LLM can automate accident information documentation in a more efficient and accurate manner. We display this potential by discussing the opportunities from the aspects of accident information extraction, imputation, and analysis.

### Accident information extraction.

The first example is accident information extraction as shown in Figure 1. Given a long narrative about the accident scene, ChatGPT can successfully extract important facts through the prompt questions. It is observed that ChatGPT can extract the keywords and fill the blank very accurately. This impressive performance benefits from large-scale mask modeling which develops a strong skill in information extraction and fill-mask. More importantly, the accident report processing with LLM could be easily parallelized with multiple GPUs, which significantly reduces data preparation time for accident analysis.

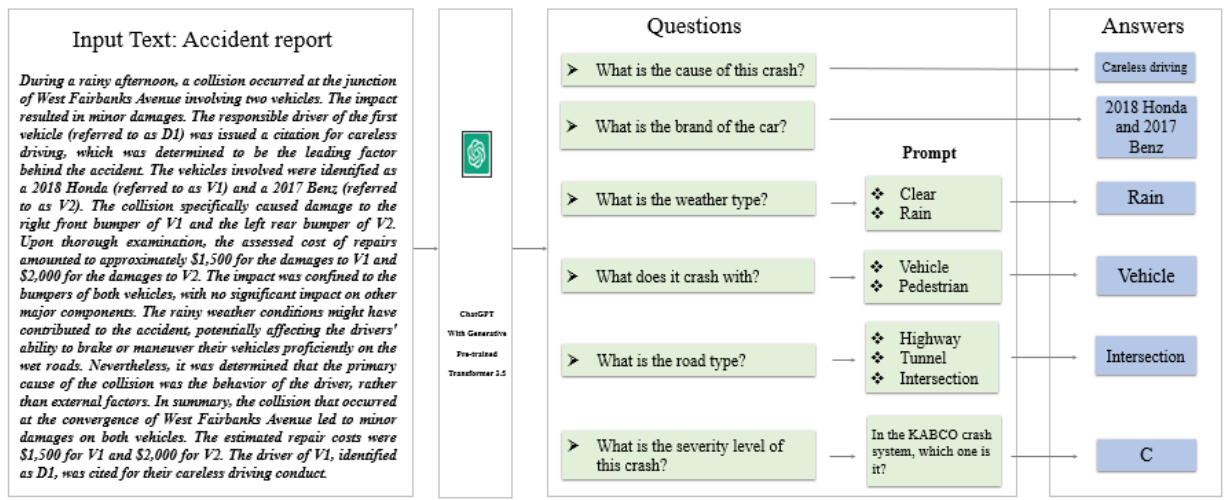

**Figure 1 Example of accident information extraction through ChatGPT.**

**Accident information imputation.**

Missing data commonly occur in the real world due to no response or observation during data collection [27, 28]. For accident analysis, when there are too many missing data, data imputation is required to reach a reliable conclusion [27]. Since accident reports are hyperdimensional textual data with semantic information, basic interpolation techniques, like statistical modeling or mathematical inference, cannot tackle this task efficiently, but it is much easier for LLMs. An LLM can leverage semantic contexts, like scene pictures or key words, to fill missing information quickly, and its reliability is also adequate. Figure 2 shows the instance the fault identification with ChatGPT. Given the brief accident news from social media, ChatGPT analyzes the accident fault and reaches the correct conclusion in the case. It is observed that the reasoning from ChatGPT is sound, and the conclusion is helpful for reference. . Sometimes, it is necessary to tackle more complex context analysis for more challenging imputation tasks. LLM can still assist in addressing this need by incorporating more diverse fine-tuning data sources, such as the California Department of Motor Vehicles (CA DMV), NHTSA, OpenStreetMap (OSM), the Open Weather API, Google street view, and official news reports, etc. By conducting LLM fine-tuning with multi-source data, a more comprehensive analysis can be undertaken to perform data imputation for the location where an accident has occurred.

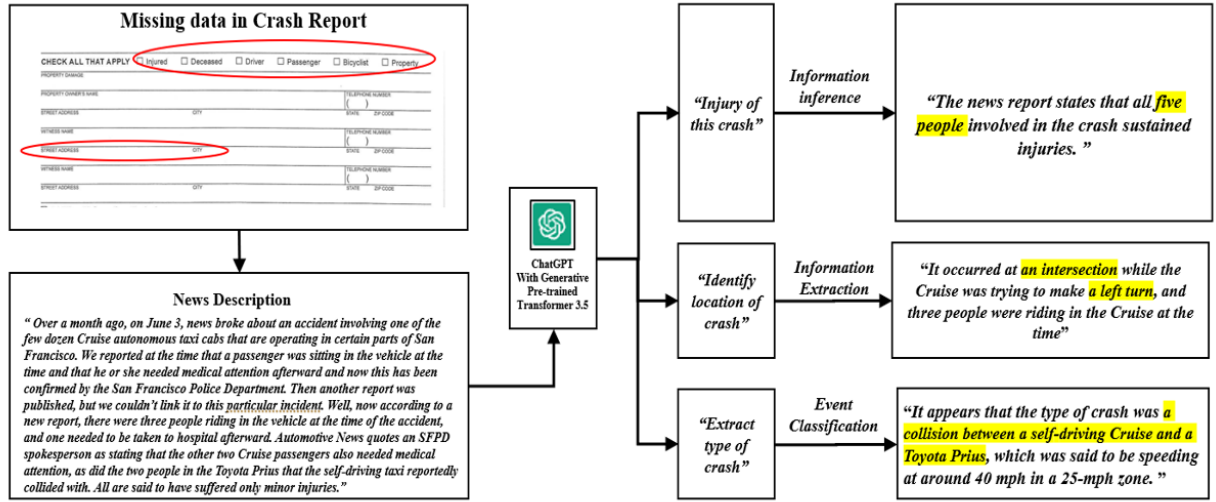

**Figure 2 Example of accident information imputation with ChatGPT.**

**Accident report analysis.**

To quantitatively analyze accident information, the narratives of accident reports must be labelled given specific research requirement. The information summarization, extraction, and classification processes demand much human effort to finish labeling. To reduce the expenditure, they can be automated with ChatGPT for data curation. We illustrate this automation in Figure 3 with an example of accident report narrative classification by ChatGPT. Given the summarization, extraction, analysis, different classification labels are acquired from the example report. These queried labels are consistent to the golden labels which are written by experts, which implies the potential of ChatGPT for accident report classification analysis. Note that the queried labels still exhibit uncertainty and need sanity check and validation by human or other models.

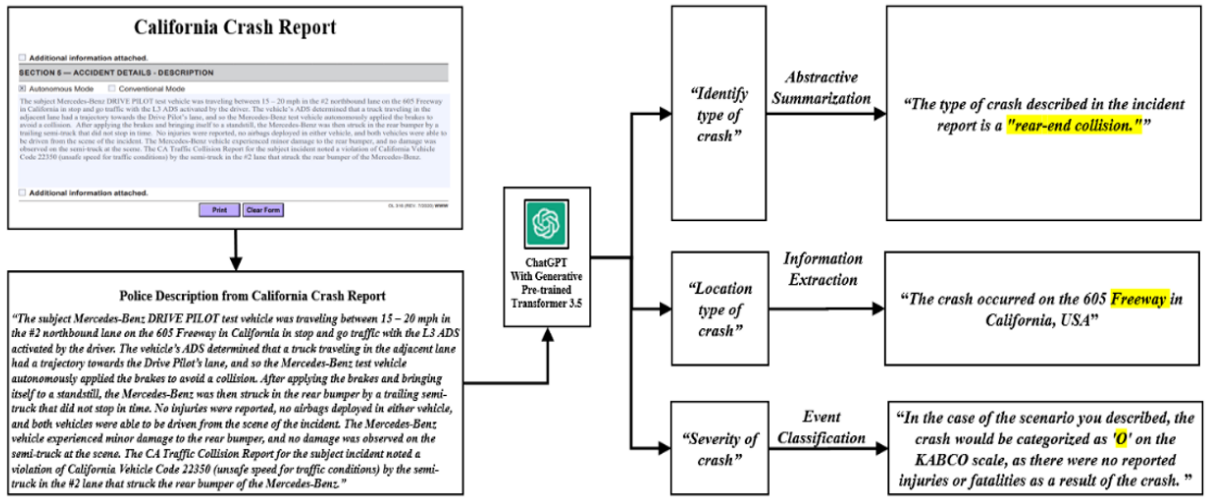

**Figure 3 Example of accident report analysis with ChatGPT.**

**Table 1 The performance of ChatGPT on accident automation tasks with exact-match evaluation.**

|  | True Positives | TPR | TF-IDF [29] |
|---|---|---|---|
| Accident Information Extraction | 18 | 0.90 | 0.0035 |
| Accident Classification | 16 | 0.80 | 0.0030 |
| Accident Data Imputation | 15 | 0.75 | 0.0016 |

We conducted a swift analysis on the performance of ChatGPT across accident automation tasks. With regard to information extraction, one question is chosen for assessment. In terms of data imputation, one blank is selected for filling. Both are evaluated by exact-match keywords. Accident classification is performed by classifying the reported accident to KABCO [30] severity levels given report description. We selected twenty accident reports to test ChatGPT and evaluate the match ratio between generated answers and human-written answers. From Table 1, we can observe that ChatGPT can accomplish these tasks with adequate accuracy. The highest accuracy reaches 0.90 while worst accuracy is 0.75 for true positive rate (TPR). This performance is almost comparable to that of the best model fine-tuned on each specific task, but ChatGPT performs direct inference without fine-tuning, i.e., zero-shot inference. We also adopted Term Frequency Inverse Document Frequency (TF-IDF) [29] of records to analyze keyword occurrences in all documents. Lower TF-IDF indicates fewer keyword occurrences and more challenging language understanding tasks. It is observed that accident data imputation with lower TF-IDF is more challenging for ChatGPT since it requires more context understanding on accident reports. We believe that the model performance on imputation tasks can be improved though more extensive training with diverse data sources would be needed. For example, researchers can incorporate additional weather data such as wind speed, rain intensity, and cloud cover, etc., from Open Weather API to improve weather condition analysis based upon the context for more effective accident weather condition imputation; more spatial data can be achieved from OSM and Google street view and assist in fine-tuning ChatGPT for better identification on road configuration.

**Traffic Data Augmentation.**

To achieve an accurate analysis model, abundant and diverse data are imperative, especially for large and complex deep neural networks. However, the observation data in the real world are limited and exhibit a long-tail distribution where some categories of data are underrepresented [31-33]. In traffic safety analysis, a long-tailed data distribution also occurs commonly [34]. For example, a blue truck is driving on the road in clear weather, which can be under-sampled in the natural images. However, it may easily fail the truck detection for computer vision models and adversely affect traffic safety decision-making. To solve this issue, data augmentation on these underrepresented classes is a straightforward and effective method [35]. An LLM can connect with an image generator, like Stable Diffusion or Segment Anything, to create synthetic data to augment these underrepresented groups. As the example in Figure 4 illustrates, ChatGPT equipped with Stable Diffusion model [19] can follow the input request to efficiently generate synthetic and true-to-life images through vision-language modeling, which can effectively improve the model for truck recognition and tracking. We perform quick assessment with Fast-RCNN [36] on object detection task with mean Average Precision (mAP). Twenty images are retrieved from Google Images and split into training and testing sets by 6-to-4 proportion. As anticipated, Fast-RCNN fine-tuned with generated images outperforms on mAP which is shown in Table 2.

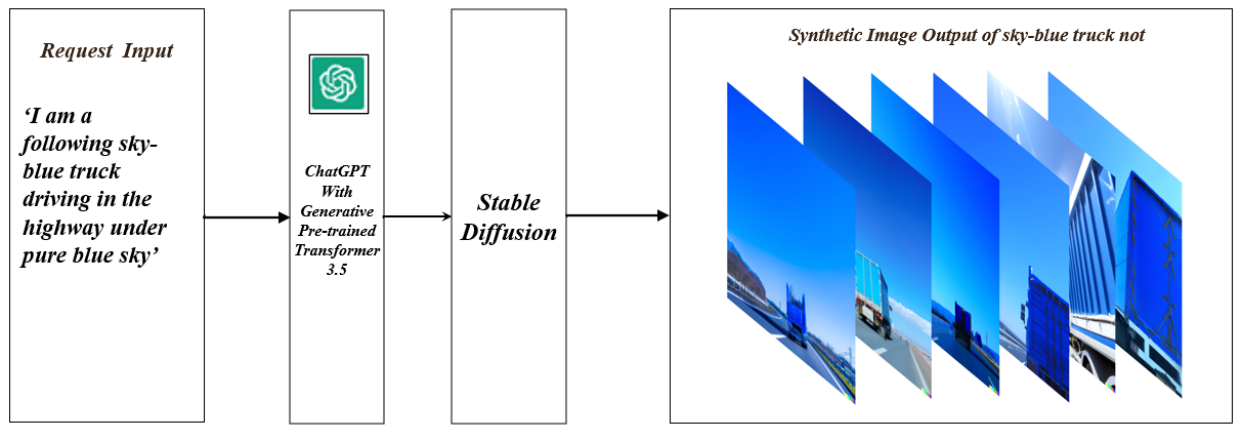

**Figure 4 Schematic of traffic scene generation with ChatGPT.** Given the prompt information, ChatGPT processes the request and collaborates with Stable Diffusion to generate data.

**Table 2 An example comparative study to assess the performance of Fast-RCNN on vehicle detection.** The generated images indicate synthetic and true-to-life images by ChatGPT and Stable Diffusion. The evaluation metric is mAP.

|  | No Fine-tuning | Fine-tuning with generated images |
| --- | --- | --- |
| Blue truck under sky | 19.2 | 34.7 |
| Green car near grass | 22.8 | 35.9 |
| White car in snow | 24.3 | 35.8 |

**Multisensory Safety Analysis.**

Multisensory analysis is essential in intelligent traffic safety [37, 38].Although existing sensor-vision integrated modeling shows some achievements, the complex analysis and decision making, like prediction with advanced reasoning, are still challenging in the event of data scarcity. For example, given only one captured image, it is difficult for computer vision models to directly predict the trajectory of the objects since there is no movement history [38]. However, an LLM can leverage the observed image features and context knowledge to infer the possible actions under specific scene settings, although the uncertainty may be higher. As Figure 5 shows, integrated with the BLIP-2 image-text model [39], ChatGPT can conduct reasonings based on image captions and predict the possible movement of this human in the scene setting. Given the reasoning, the model derives that the man will likely cross the street because he pushes the button. This analysis result can help to manage traffic lights more efficiently. When more pictures are provided, the intention of the human action can be analyzed more accurately, which enables more efficient traffic safety management [22]. However, please note that there are certain risks associated with integrating these data unless reliable modeling or human expert curation is employed.

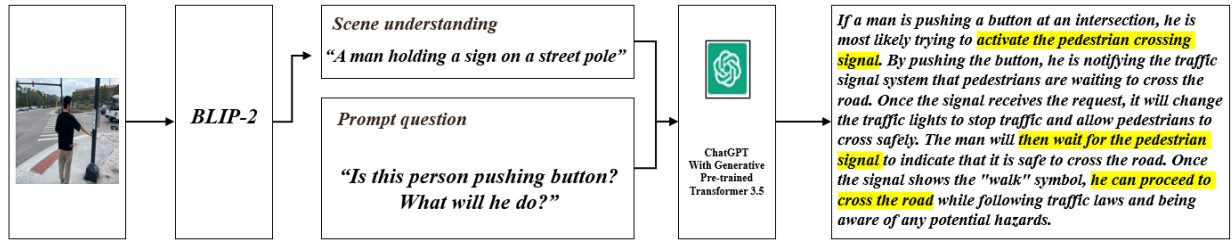

**Figure 5 Schematic of multisensory prediction and analysis with ChatGPT**. Given the scene understanding from BLIP-2 and prompt questions, ChatGPT generates a narrative for action inference, which can be extracted for action sequence analysis. Note that efficient and reliable text extraction is required to ensure narrative extraction quality.

Another important application is mobile data collection and integration. With rapid development of edge computing, more mobile devices, like cell phones, can process more complex and diverse data by communicating with cloud servers [40-42].Intelligent traffic safety systems can leverage this technology to further improve data collection efficiency. Figure 6 illustrates a demonstration of automated accident report generation with smartphones. A smartphone is usually equipped with various sensors and devices that can collect different types of data, including image/video (camera), audio (microphone), location (GPS), and text (keyboard). An accident scene can be captured from multiple sources through a smartphone by leveraging these advanced features. Then, multisource data can be transcribed to texts or converted to a unified format, like embedding, and feed these data to an LLM, such as ChatGPT, for integrated data modeling. Finally, an accident report can be generated through text summarization and generation. This process may involve visual question answering, speech recognition, natural language undemanding, and natural language generation.

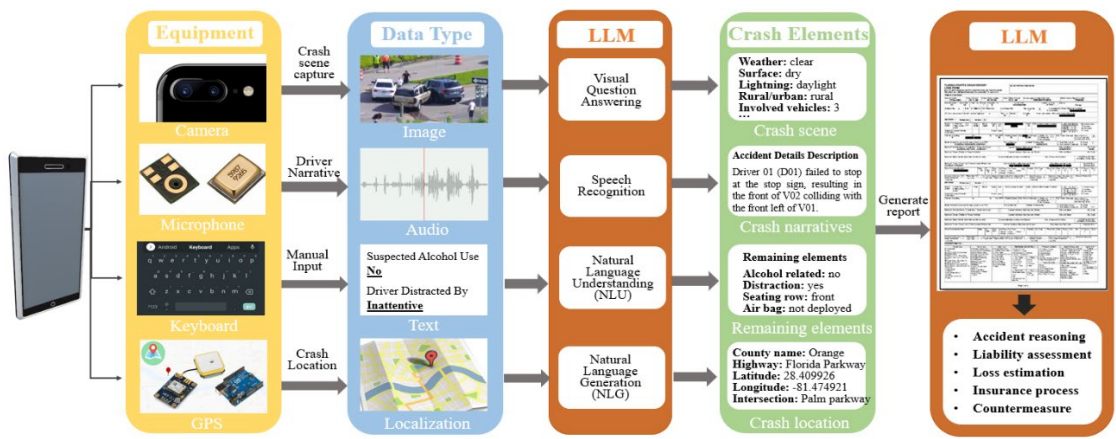

**Figure 6 A schematic of accident report generation with multisensory analysis via ChatGPT. Multisensory data are collected from different devices.**

## THE POTENTIAL RISKS OF LLM IN TRAFFIC SAFETY APPLICATIONS

Although LLMs show potential in successful applications for traffic safety management, several critical issues and limitations may prevent the wide deployment of these applications as a customer service in the real world. Since the current variants of LLM are still rooted in data-driven learning, this problem-solving paradigm still has some important limitations. When defective data intervenes the model training, model bias will probably occur which harms user experiences and causes severe issues like discrimination. Another critical issue is data privacy. To ensure the best performance of LLM, numerous data are shared for communication, which increases the risk of data breaching. Meanwhile, the application integrated with LLM can be vulnerable since deep neural networks are sensitive to input perturbations due to inverse modeling paradigm. Moreover, artificial hallucination, an important issue of LLM, may mislead decision making because the prepared dataset, even at a large scale, cannot encompass all diverse observations, which yields inaccurate model inference on unseen data. Over these typical questions, we further investigate the problems and possible solutions.

### Model Bias.

The collected data, even large-scale data, hardly represents the whole real world, which affects the reliability of data-driven modeling, like deep learning. Deep learning relies on large amount of training data and these data not only benefit specific tasks, but can also manipulate the modeling with inductive bias [43]. In NLP, there are some important problems owing to inductive bias from textual data, such as gender-bias [44] and racism [45]. These issues, which may cause partial understanding of human behaviors, can also yield inaccurate modeling and impede application of LLM for traffic safety analysis [44, 46]. For example, the collected accident documents or traffic data are imbalanced over all groups [47], it can yield a biased model with less attention to the safety of minority groups, which places their lives at risk. These critical issues always concern the researchers and practitioners on LLMs [48, 49]. Various research efforts in the fields of traffic safety record documentation and text mining are dedicated to developing solutions that detect and mitigate discrimination within collected datasets [50, 51]. For traffic safety studies, we maintain that this issue requires more serious consideration over the data preparation to ensure a reliable traffic safety application. The potential approaches can be bias auditing, ensemble method, and fair representation learning. For bias auditing, we recommend human-in-the-loop [52] for misprediction correction and model fairness improvement. Also, ensemble method can be used to alleviate partiality problems by aggregating diverse results from multiple models. More importantly, textual data need debiasing by neuralization, like gender-neutral language [53], for fair representation learning.

### Data Privacy.

The striking success of LLM is achieved through training with voluminous data which raises a range of concerns about data privacy management. To achieve large-scale deployment of large models, frequent data communication is inevitable between server and edge devices, and thus, efficient data transmission is usually incorporated into applications [54]. These efficient practices unleash the power of LLMs, but also take more risk on data breach. Therefore, privacy preservation issue is a dominant feature in the minds of NLP researchers and practitioners and a range of initial works have been conducted to tackle this problem [55, 56]. To intelligent traffic safety, data privacy is also an important concern when LLM is applied to solve issues. For instance, real-time analysis and prediction requires intensive data streaming which may cause sensitive data exposure due to high vulnerability to cyberattack [57, 58]. To protect sensitive traffic data from threat, the application equipped with LLMs must be improved with secured privacy protection. Since intelligent traffic safety applications involve edge devices, we suggest federated learning with edge computing [59] as one of possible solutions for intelligent traffic safety systems. The personal devices conduct edge computing to fine-tune a large foundation model with local data shards for customized solution, and at the same time to upload the local model copy to cloud data center. The communication pattern reduces the threats of data breaching and improves data privacy protection. Its implementation may need more efficient network communication support, but we consider this strategy as a more practical solution because of its stronger privacy-preserving [60]. Furthermore, federated learning enables local application fine-tuning with data minimization [61]. Since local edge devices have specific functionalities for traffic safety assessment or prediction, it can alleviate the demand for a huge set of data and reduce the potential exposure of sensitive information.

**Model Vulnerability.**

Nowadays, more applications harness the potential of large deep neural networks to solve complex problems. Although these applications effectively improve our daily life, they are very vulnerable since the robustness of deep neural networks is still an important issue due to its inverse modeling paradigm [62-64]. An imperceptible perturbation on input data can cause significant change in model output and severe prediction degradation, which is caused by weak interpretable modeling on feature learning of deep neural network [65, 66]. To address this issue, adversarial machine learning becomes a more important field to understand how to deteriorate deep model performance and improve neural network robustness [65-67], where some approaches are developed such as robust training [68],ensemble learning [69], and threat modeling [70]. Undoubtedly, this issue is of significant importance to stakeholders in the domain of intelligent traffic safety [71].After considering several existing solutions, we suggest threat modeling for traffic safety application improvement. The main idea is to develop a threat model that resembles the real application and leverage it to conduct extensive and thorough vulnerability analysis. The researchers can incorporate adversarial training into this paradigm for model robustness improvement. Also, continual learning [72] is recommended to enhance the recognition of the new features, which relies on the model maintenance with regular security assessment and upgrading with fine-tuning. Furthermore, ensemble learning [73] can equip threat model study to derive a stronger solution by aggregation of diverse robust models. Since it may increase storage space or inference time, some efficiency improvement approaches can be integrated, like knowledge distillation.

**Artificial Hallucination.**

Large deep neural networks rely on substantial amounts of training data, but a prevailing concern is the constrained accessibility to big data. Data augmentation is widely used to address this data limitation problem and improve the generalization of large models in computer vision [74], signal processing [75], and NLP [76]. This prevalent technique leverages synthetic data to render the model hallucinate nonexistent data in training set. Although it effectively enhances feature recognition on unseen real data, it can cause the problem of artificial hallucination [77, 78]. This problem from data fabrication becomes more severe in LLM, where an LLM offers a plausible but completely incorrect answer [79, 80]. Undoubtedly, this issue will place the LLM applications that serve intelligent traffic safety into risk. For example, the generated accident report analysis from ChatGPT in Figure 7 mentions a "semi-truck"; however, this word has nothing to do with the accident incident report. Also, the conclusion that "the driver violated California vehicle code 22350" is hasty without a solid foundation.

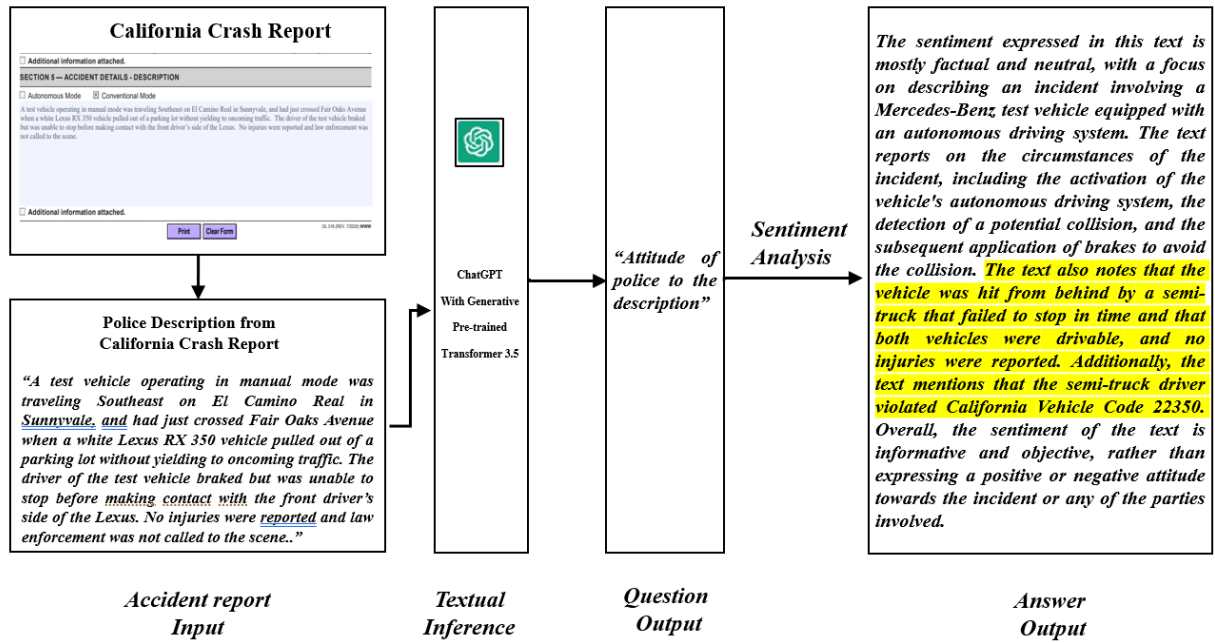

**Figure 7 An example of artificial hallucination in accident report analysis.** The highlight generated text shows a "semi-truck", which has nothing to do with the accident incident report.

To this problem, we advocate that training data quality is a priority to secure the reliability of LLM and alleviate artificial hallucination [81]. For traffic safety LLM, accurate data are critical to a reliable prediction model. The practitioners need careful consideration of data quality and ensure strong data distribution fidelity to eliminate potential bias. When data augmentation is conducted, further data scrutinization is imperative to avoid distribution shift and artificial hallucinations on unseen scenarios. For instance, the generated texts from LLM may need sanity testing by human experts to ensure their accuracy and completeness, while it may incur more labor cost.

## AFTER CHATGPT

Despite its merits and potential risks, LLM provides intelligent traffic safety a new solution by foundation model training and the power of large-scale comprehensive data. We maintain that LLMs like ChatGPT just embark on a new era of large-scale representation learning and the learning paradigm can empower current intelligent traffic safety system by integrating the data collected from an expanding range of sensors in Figure 8(a). We lay out an illustration of the interaction between textual data encoded by LLM and other different data modalities, including image and signal, for intelligent traffic safety problem solving in Figure 8(b). They exhibit distinct features and functionality for problem solving and can mutually enhance the productivity for real applications. However, it is a challenging task to incorporate all information, particularly domain-specific expertise, into a single model. Even pre-trained LLMs like ChatGPT can handle multiple tasks simultaneously, but their performance on highly specialized tasks within specific domains is still moderately satisfactory. One of the straightforward methods to approach it is domain-specific fine-tuning. It is a favorable choice to leverage domain knowledge for fine-tuning an LLM to achieve improvement. Thanks to the LLaMA family [83], the cost of fine-tuning is notably diminished for regular users. For example, LLaMA-adapter [84] introduces merely 1.2 million trainable parameters to the frozen LLaMA-7B model and requires less than an hour for fine-tuning using 8 A100 GPUs. Qlora leverages quantization on model parameter representation to effectively reduce the fine-tuning cost to single 48GB GPU [85]. Despite computation efficiency improvement in LLM fine-tuning, it is important to give further thought to data availability and quality, which may cause weak generalization or artificial hallucination.

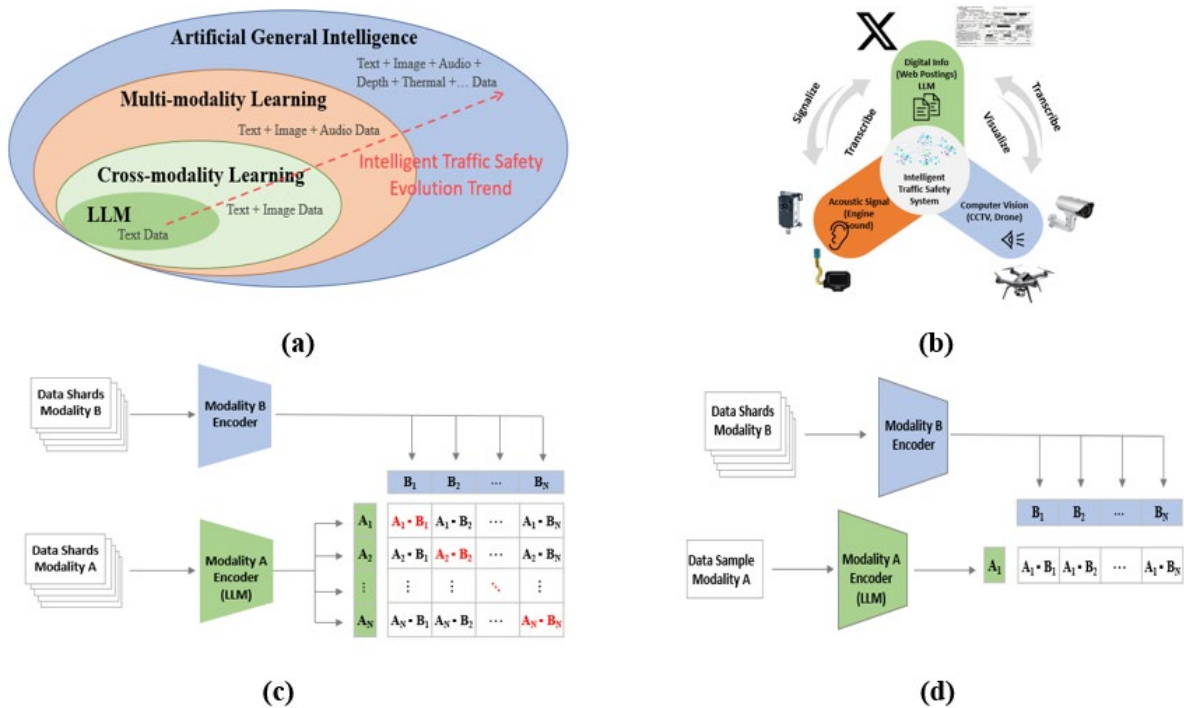

**Figure 8  The proposition for the advancement of intelligent traffic safety after ChatGPT. (a)** Trend in the evolution of intelligent traffic safety. **(b)** The interaction between the textual data and other modality data through cross-modality modeling with domain-specific model. **(c)** Cross-modality learning between textual encoder of LLM and the encoder of another modality. **(d)** Cross-modality inference with textual encoder of LLM to achieve relevant data from another modality.

In addition to LLM fine-tuning, we argue that it is possible to integrate multi-modalities to equip our intelligent traffic safety system with "eyes and ears" to sense, interact, and respond to the physical world, though successful modality fusion is not an easy task. We think the paradigm proposed by CLIP [82] could be a tentative solution. It not only effectively integrates two sources in a simple manner, but also leverages unlabeled data which saved data curation labors. Derived from CLIP, any modality can be correlated with textual representation from LLM encoder by contrastive representation learning through correlation optimization in Figure 8(c) and prompt testing in Figure 8(d). Figure 8(c) depicts cross-modality learning formulated with contrastive learning given the textual representation from LLM encoder and another data source of modality B, where the correlations between red matched data pairs are enhanced while the correlations of other mismatched pairs are weakened by penalty learning. Figure 8(d) shows contrastive cross-modal learning inference process with one observation from textual representation of LLM encoder, where the modality B for fusion provides test data shards, and the most matched data is searched out by correlation information. After this modality fusion, once the observed textual data are obtained, the relevant data from another modality can be retrieved with the prompt information. Although the quality of retrieved data may not be as good as true observations, they can perform axillary decision making to boost intelligent traffic safety system.

From our perspective, the research on next intelligent traffic safety system ought to focus on: (1) more efficiently projecting the data from different modalities into a large and comprehensive space; (2) inclusively represent the data from different modalities; (3) exploration in multi-sensory problem formation with specific metrics given research problem or scales; (4) further examination on model bias, data privacy, model vulnerability, and application security and efficiency.

## CONCLUSIONS

We shared our perspectives on the impact of LLM on intelligent traffic safety research. This emerging technology is a double-edged sword, creating more efficient solutions and more unpredictable challenges. For the improvement in traffic safety solution, we showcase the potential of LLM in automating accident reports, including accident information extraction, imputation, and analysis, as well as augmenting traffic data, and conducting multisensory safety analysis. The discussed applications show the new opportunities to facilitate traffic safety decision-making and improve traffic safety services. Meanwhile, the implementation of these applications can bring new threats to traffic safety analysis. The threats are including but not limited to model bias, data privacy, model vulnerability, and artificial hallucination. We argue that it is critical to further explore and understand the features of the LLMs like ChatGPT from more aspects before we apply them to more benefit intelligent traffic safety. Furthermore, utilizing a large language model as a foundation, we can create an even more potent model through multi-modality learning, which equips intelligent traffic safety system with multiple sensors to enable it to emulate human decision-making. This provides more unexplored territory available for thorough and comprehensive research investigation.

## AUTHOR CONTRIBUTIONS

The authors confirm contribution to the paper as follows: Dr. Ou Zheng, Dr. Mohamed Abdel-Aty, and Dr. Dongdong Wang conceived the study. Dr. Ou Zheng, Dr. Mohamed Abdel-Aty, Dr. Dongdong Wang, Zijin Wang, and Shengxuan Ding wrote the manuscript. Zijin Wang and Shengxuan Ding implemented the case studies and carried out the experiments. All authors reviewed the results and approved the final version of the manuscript.